\documentclass[letterpaper, 10 pt, conference]{ieeeconf}  

\IEEEoverridecommandlockouts                              

\usepackage{color}
\usepackage{graphicx}
\usepackage{amssymb}
\usepackage{amsmath}
\usepackage{threeparttable}
\usepackage{algorithm}
\usepackage{algorithmic}
\usepackage{cite}
\usepackage{url}
\usepackage{bm}
\usepackage{footnote}
\usepackage{float}
\usepackage{booktabs} 
\usepackage{multirow}
\usepackage{lipsum}

\title{\LARGE \bf Probabilistic Trajectory Prediction for Autonomous Vehicles with Attentive Recurrent Neural Process
}

\author{Jiacheng Zhu, Shenghao Qin, Wenshuo Wang, {\it Member, IEEE}, and Ding Zhao
\thanks{The first two authors, J. Zhu and S. Qin, make equal contributions. (Corresponding Authors: Wenshuo Wang and Ding Zhao.)}
\thanks{J. Zhu, S. Qin, W. Wang, and D. Zhao are with the Department of Mechanical Engineering, Carnegie Mellon University (CMU), Pittsburgh, PA 15213, USA. {\tt\small jzhu4@andrew.cmu.edu},
{\tt\small shqin16@fudan.edu.cn},
{\tt\small wsbit@gmail.com}, 
{\tt\small dingzhao@cmu.edu}}%
}

\begin{document}

\maketitle
\thispagestyle{empty}
\pagestyle{empty}

\begin{abstract}

Predicting surrounding vehicle behaviors are critical to autonomous vehicles when negotiating in multi-vehicle interaction scenarios. Most existing approaches require tedious training process with large amounts of data and may fail to capture the propagating uncertainty in interaction behaviors. The multi-vehicle behaviors are assumed to be generated from a stochastic process. This paper proposes an attentive recurrent neural process (ARNP) approach to overcome the above limitations, which uses a neural process (NP) to learn a distribution of multi-vehicle interaction behavior. Our proposed model inherits the flexibility of neural networks while maintaining Bayesian probabilistic characteristics. Constructed by incorporating NPs with recurrent neural networks (RNNs), the ARNP model predicts the distribution of a target vehicle trajectory conditioned on the observed long-term sequential data of all surrounding vehicles. This approach is verified by learning and predicting lane-changing trajectories in complex traffic scenarios. Experimental results demonstrate that our proposed method outperforms previous counterparts in terms of accuracy and uncertainty expressiveness. Moreover, the meta-learning instinct of NPs enables our proposed ARNP model to capture global information of all observations, thereby being able to adapt to new targets efficiently.


\end{abstract}

\section{Introduction}

\begin{figure}[t]
      \centering
      \includegraphics[width=1.0\linewidth]{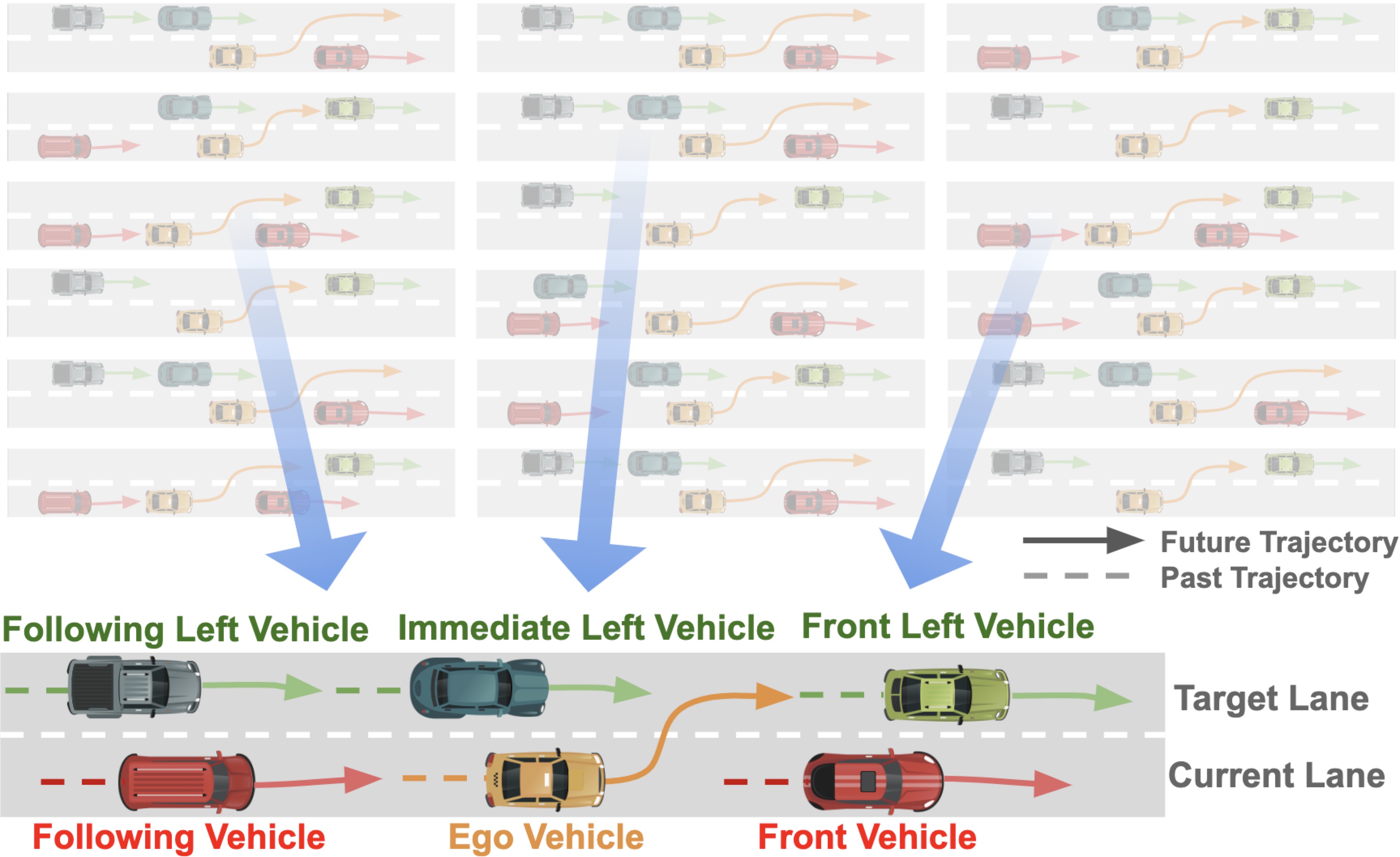}
      \caption{ The are numberless distinct left lane-changing scenarios, with varying initial conditions such as the velocities, number of vehicles, and relative position of surrounding vehicles. The button of the figure shows one typical left lane-changing scenario: The orange \textit{Ego Vehicle} is the target vehicle, the trajectory of which is estimated according to its surrounding vehicles such as the leading \textit{Front Vehicle}, the \textit{Following Vehicle}, the closest \textit{Immediate Left Vehicle}, its {\it Front Left Vehicle}, and its following \textit{Following Left Vehicle}.}
      \label{lane_changing_scenario}
\end{figure}

As autonomous vehicles (AVs) technology gradually progresses from algorithms and simulation to real-world testing and on-road services \cite{bloomberg_waymo_deploy}, recognizing and understanding complicated mixed urban traffic scenarios has become the critical bottleneck for AVs to operate reliably \cite{litman2017_AV_imple_pred}. Understanding and predicting the trajectory of human-driven vehicles through interactively considering the future behaviors can benefit the safe planning and efficient decision-making processes of AVs. Take a typical lane-changing scenario \cite{Chiyu2019_cnp_traj} as example (see Fig. \ref{lane_changing_scenario}), the human driver (labeled as Ego Vehicle) is going to change into a nearby target lane from its current lane. The driver would make a decision, select a suitable cutting-in gap, and generate the desired trajectory to follow according to the recent behaviors and movements as well as the prediction of the surrounding vehicles. Thus, precise prediction of this driver behavior allows the surrounding vehicles (such as following left vehicle) to make a proper decision such as decelerating and leaving a space for the ego vehicle. The prediction of vehicles mainly covers two aspects: intention/maneuver/behavior prediction and motion/trajectory prediction \cite{hu2018probabilistic}. The former one mainly emphasizes on high-level decision-making outputs such as determination of turning left or keeping lanes, while the latter one would generate an expected trajectory over time. Most research has provided probabilistic solutions to the high-level decision making such as in \cite{ding2019predicting,cui2019multimodal}. In this paper, our focus is on the motion prediction to generate time-profiled trajectories of target vehicles over continuous space.

Human drivers can navigate in complex scenarios thanks to the their powerful ability to make a reasoning and prediction of the vehicle behavior by efficiently exploiting and actively exploring surrounding information, but it is still challenging for AVs to make a human-like prediction of vehicle trajectory to effectively navigate under the same circumstance. In order to predict a target vehicle's (manipulated by a human) trajectory precisely, the mutual interactions between the target vehicle and all the surrounding vehicles should be taken into consideration. Generally, it is inadequate to ignore the uncertainty for motion/trajectory prediction using a deterministic model in complex driving scenarios \cite{ding2019predicting}. This is because the real-world human driver behaviors share similar behavioral elements \cite{wang2019drivingstyle}, but vary a lot actually, as shown in Fig. \ref{lane_changing_scenario}. This brings three critical challenges in predicting vehicle trajectories: (1) Learning sufficient representations that consider mutual interaction information among all related surrounding vehicles; (2) Predicting trajectories with the uncertainty which is functionally characterized by explicit distributions; and (3) Developing a generalized model that can summarize and learn the underlying knowledge from limited size of data, thus can adapt to new scenarios efficiently.

In order to overcome the aforementioned challenges, we propose a model by integrating a neural process involved attention mechanisms with a recurrent framework, called attentive recurrent neural process (ARNP), to comprehensively model the distribution over functions of observed vehicle trajectories in complicated traffic scenarios. By assuming that the observed trajectories are generated from multiple realizations of a stochastic process, our proposed model first comprehensively captures the sequential interaction behavior through a RNN framework, and then the predictive trajectory is obtained by estimating a distribution of the mapping from surrounding vehicle information to corresponding target vehicle trajectory conditioned on the history input-output pairs.  


The remainder of this paper is organized as follows. Section II reviews the related work of vehicle trajectory prediction. Section III introduces our proposed Attentive Recurrent Neural Process for trajectory prediction. Section IV presents the experiments and result analysis. Section V summarizes this current work and discusses future work.

\section{Related Works}

\begin{figure*}[t!]
      \centering
      \includegraphics[width=1.0\linewidth]{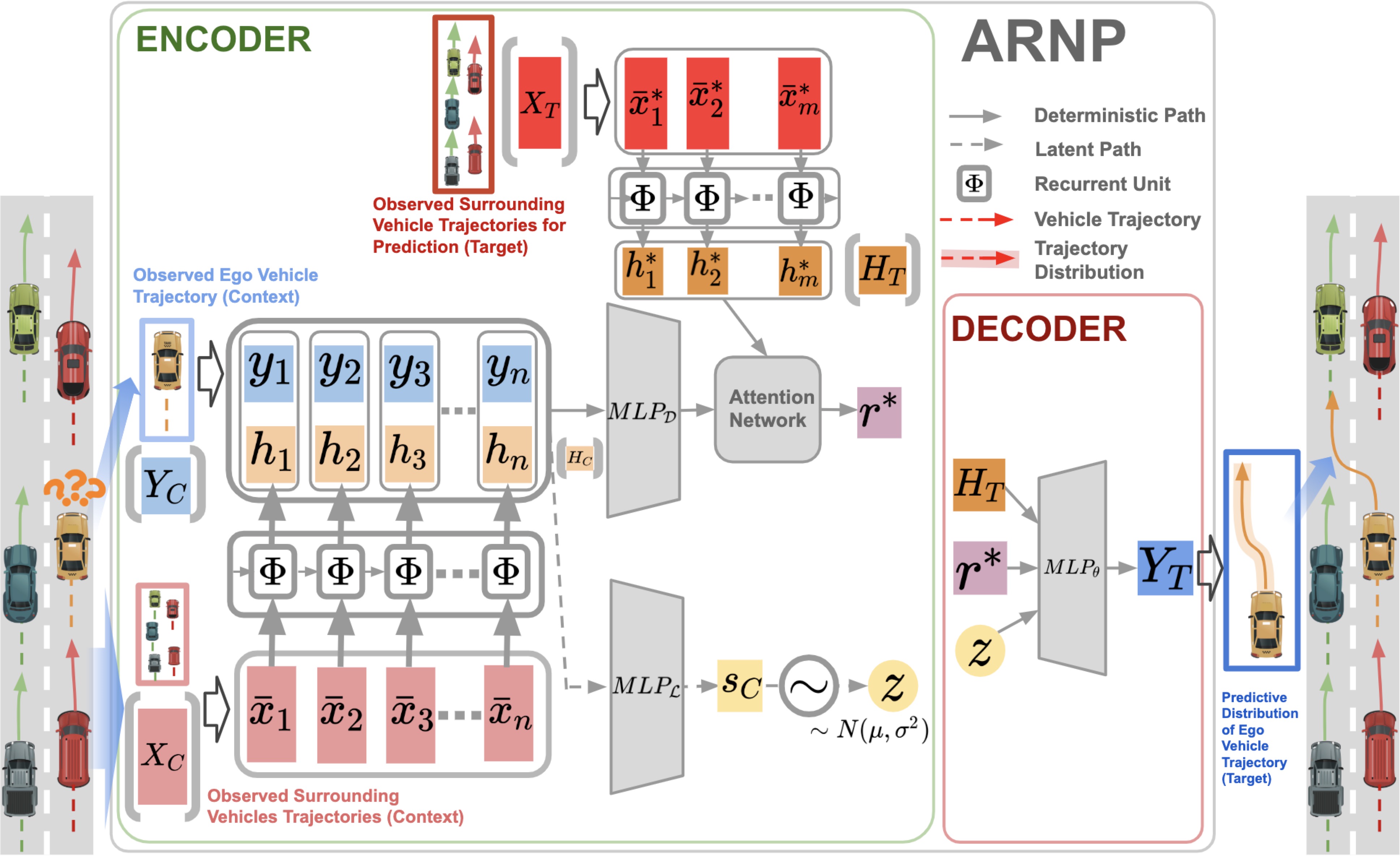}
      \caption{The framework of the proposed Attentive Recurrent Neural Process (ARNP) model to predict the target vehicle trajectory in a lane-changing scenario. The observed surrounding vehicle trajectories $X_C$ are first passed into a recurrent neural network to obtain a latent representation $H_C$.
      The global structure of the stochastic realization is captured by $z$ in the latent path, where as the fine-grained local structure along with the relevant information of \textit{target} inputs and observed \textit{context} attended. The likelihood is referred to as the decoder. The ARNP model allows one to model the vehicle trajectory as different realizations of a stochastic process.}
      \label{framework}
\end{figure*}

Vehicle trajectory prediction may occur in different scenarios such as curve-tracking, lane-changing, and intersection negotiation \cite{huang2019uncertainty}. In this research, our focus is mainly on the lane-changing behavior, which is one of most changing tasks in self-driving applications. Many research regarding trajectory prediction and generation are proposed for AVs in the past decades. These approaches are ranging from optimization with a deterministic model to stochastic inference with a probabilistic model. 

For optimization-based methods, a specific controller such as model predictive control (MPC) \cite{LCL2017convex_mpc} is usually applied for the cooperative trajectory planning, in which the controller can be formed as a convex optimization over its manifold. However, this optimization-based method with prediction capability needs a predictive engine for initial estimation of other vehicles. An alternative for trajectory prediction is using the empirically-build model such as the Intelligent Driver Model (IDM) for car-following trajectory prediction \cite{treiber2000_IDM}. In this model, the concerned factors are usually selected and defined according to prior knowledge \cite{hou2013_emp_lanec,yang2016_emp_lanc}.  This kind of models assumes that all the vehicles follow an identical IDM, and optimize vehicle reactions in interaction-required scenarios. However, this assumption can not fit all vehicles, thus being limited to real-world traffic scenarios. 

In order to overcome the limitations of optimization and empirical-based approaches, probabilistic methods are developed for modeling cooperative driving and lane-changing behaviors by considering uncertainty. For instance, a global path planner developed by Stanford Junior \cite{montemerlo2009junior_stanford} was developed to predict lane-changing trajectories by optimizing a variant Bellman equation under the Markov Decision Process (MDP). In this approach, the lane-changing behavior of the ego vehicle was treated as a penalty term in the cumulative cost function; however, it can not be applied for other near traffic participants (i.e., the surrounding vehicles). One of the key reasons is that not all the underlying state can be directly observed by the agent. To overcome this issue, the partially observable MDP (POMDP) \cite{szepesvari2010_pomdp} were utilized \cite{ulbrich2013_pomdp_lanechange,wei2011_pomdp_lanechange} to model lane-changing behavior with a real-time belief space search algorithm. Nevertheless, the state and action spaces are non-continuous. Bai, et al. \cite{bai2014_conti_pomdp} relaxed this limitation by proposing a continuous-state POMDP using a belief tree. An online and approximate solver \cite{seiler2015online_conti_POMDP} has been developed for a continuous action, but solely tested in toy examples. In addition, the reproducing kernel Hilbert space (RKHS) \cite{Chiyu2018_itsc_rkhs} was applied to approach a nonparametric regression task for predicting discrete and continuous trajectories. 


The machine learning-based approaches have been widely used for learning and predicting vehicle trajectories due to their powerful capability to deal with nonlinearity and uncertainty of complex traffic scenarios. For example, the $K$-nearest neighbors (KNN) methods is applied in a lane-change scenario \cite{yao2013KNN_lane} to specific and generate trajectories. 
Besides, reinforcement learning (RL) is also used to model interaction behaviors among vehicles at intersections \cite{qiao2018_RL_intersection}. Further, \cite{sadigh2016information_IRL} established the transition models for Inverse Reinforcement Learning (IRL), but their methods are limited to specific scenarios. \cite{ding2018_encounter} proposed a Multi-vehicle Trajectory Generator (MTG) based on Beta-Variational Auto-Encoder ($\beta$-VAE) for trajectory data augmentation. Generative Adversarial Networks (GANs) \cite{goodfellow2014_GAN} has shown its capability in imitation and estimation \cite{kuefler2017imitating_gan} of single vehicle trajectories. A generative neural system \cite{li2019conditional_lijiachen} is proposed for generating trajectories hypotheses instead of probabilities. \cite{Chiyu2019_cnp_traj} proposed a Recurrent Meta Induction Neural Network (RMIN) by utilizing a condition neural process as generator, whereas the explicit probability is still not given.


\section{Attentive Recurrent Neural Process}
\label{section3}
In this paper, we develop a comprehensive method to learn and predict the target vehicle trajectory in a lane-changing scenario according to the behaviors of all surrounding vehicles. However, there are many lane-changing behaviors which share substantial similarity but vary slightly caused by uncertainty. Therefore, it is reasonable to treat different lane-changing behaviors as multiple realizations of a stochastic process, as shown in Fig. \ref{framework}.\par

\subsection{ARNP Model}

\begin{figure}[t]
      \centering
      \includegraphics[width=0.7\linewidth]{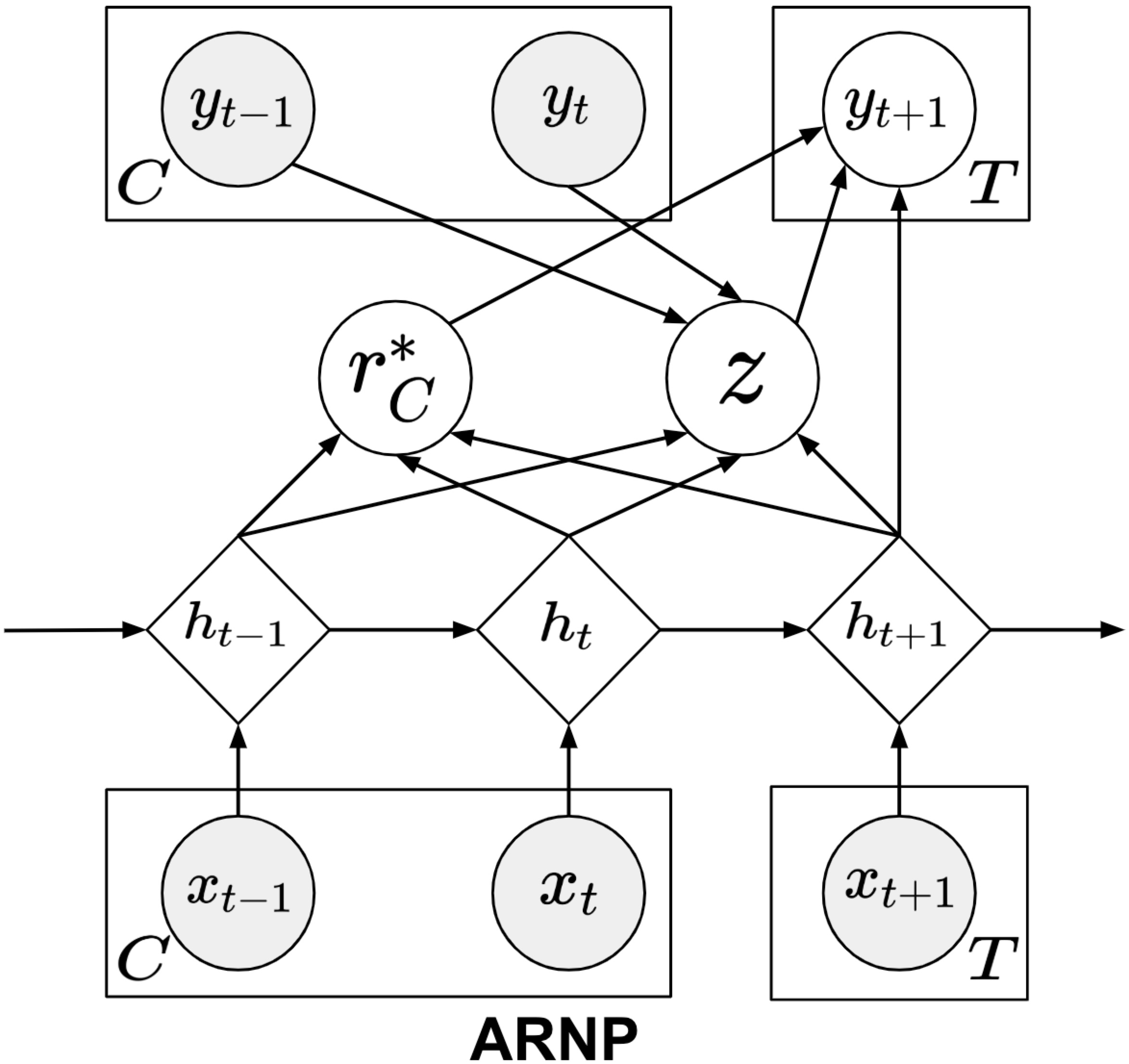}
      \caption{The graphical illustration of the ARNP model.}
      \label{figurelabel}
\end{figure}

Let the trajectory of the target vehicle be a collection of real-value vectors $\boldsymbol{y} = \{ \boldsymbol{y}_t\}_{t=1}^n$ and $\boldsymbol{y}_t \in \mathbb{R}^d$, where $n$ is the length of trajectory. The surrounding vehicle trajectories, which is the sequence of positions, are formally depicted as vectors of measurements $\boldsymbol{\bar{x}_1} = [\boldsymbol{x^1}] \text{ ,  }  \boldsymbol{\bar{x}_2} = [\boldsymbol{x^1}, \boldsymbol{x^2}], \dots, \boldsymbol{\bar{x}_n} = [\boldsymbol{x^1}, \boldsymbol{x^2}, \dots,\boldsymbol{x^n}]$, and are indexed by time and would be of growing lengths. Assuming only the most recent $L$ steps of surrounding vehicle positions are predictive of the target vehicle trajectory, and let $\boldsymbol{\bar{X}} = \{ {\boldsymbol{\bar{x}_t}} \}^n_{t=1}$ be a collection sequences $\boldsymbol{\bar{x}_i} = [\boldsymbol{x^{i-L+1}}, \boldsymbol{x^{i-L+2}},...,\boldsymbol{x^{i-L+L}}]$ with corresponding length $L$, where $\boldsymbol{x}^i \in \mathcal{X}$. Hence, for each distinct lane-changing behavior, the exact relation between surrounding vehicle behavior and target vehicle trajectory is described as a \textit{function}, $f : \mathcal{X} \mapsto \mathbb{R}^d$. \par

In order to capture the sequential information from the surrounding vehicle trajectories. Here an input sequence $\boldsymbol{\bar{x}_t}$ is converted to a latent representation $\boldsymbol{h}^L_i \in \mathcal{H}$ by recurrent neural networks.
\begin{equation}
\begin{aligned}
\boldsymbol{h}^i_t = \Phi(\boldsymbol{h}^{i-1}_t, \boldsymbol{x}^{t-L+i})+\delta^t \text{ ,  }i=1,...,L
\end{aligned}
\label{eq1_rnn}
\end{equation}
where the recurrent model $\Phi : \mathcal{X} \mapsto \mathcal{H}$ is the transformation which maps $\{\boldsymbol{\bar{x}}_t \}^n_{t=1}$ to $\{\boldsymbol{h}^L_t \}^n_{t=1}$ so that the latent representation is obtained. \par
Subsequently, based on aforementioned stochastic process assumption, the underlying dynamics of vehicle trajectory can be revealed by learning the distribution of regression function in latent space, on which one may conditioned on an arbitrary number of observed \textit{contexts} $(H_C, Y_C):= (h_t,\boldsymbol{y}_t)_{t \in C} = (\Phi(\boldsymbol{\bar{x}_t}), y_t)_{t \in T}$ to model an arbitrary number of \textit{targets} $(H_T, Y_T):= (h_t,\boldsymbol{y}_t)_{t \in T}= (\Phi(\boldsymbol{\bar{x}_t})_{t \in T}, y_t)$, additional the original input-output pair for function $f$ as given as $(X_T, Y_T):= (\boldsymbol{\bar{x}_t}, \boldsymbol{y}_t)_{t \in T}$ and $(X_C, Y_C):= (\boldsymbol{\bar{x}_t}, \boldsymbol{y}_t)_{t \in C}$. The generative process of ARNP \cite{Qin2019attentive_ANPRNN,louizos2019functional_FNP,willi2019recurrent} (as shown in Fig. \ref{figurelabel}) is given as 
\begin{equation}
\begin{aligned}
p(Y_T | H_T, H_C, Y_C):= \int p(Y_T|H_T,r^*_C, z)q(z|s_C) dz
\end{aligned}
\label{eq2_np_gen}
\end{equation}
where $z$ is the global latent variable describing the uncertainty in the predictions of $Y_T$ conditioned on observation in transformed space $(H_C, Y_C)$, and is modeled by a factorized Gaussian parameterized by $S_C:=s(H_C, Y_C)$, in which $s$ is an encoder representing the \textit{contexts} $(H_C,Y_C)$. Meanwhile, $r^*_C$ is a deterministic attention function $r^*_C:= r^*(H_C,Y_C,H_T)$ that forms the relevant information among $contexts$ and $targets$ via attention mechanisms\cite{kim2019attentive_ANP}.



\subsection{Learning and Inference of ARNP} 

Remind that the input sequences $\{{\boldsymbol{\bar{x}_t}} \}^n_{t=1} \in \mathcal{X}^n$ are mapped into latent representations $\{h_t \}^n_{t=1} \in \mathcal{H}^n$ using recurrent cells, and we denote $H_C = \Phi(\boldsymbol{\bar{x}_t})_{t \in C}$ and $H_T = \Phi(\boldsymbol{\bar{x}_t})_{t \in T}$. The Evidence Lower Bound (ELBO) can be derived as

\begin{equation}
\begin{aligned}
\log p(Y_T | X_T, X_C, Y_C) &\geq \\
\mathbb{E}_{q(z|H_T,H_C,Y_C)}&[\log p(Y_T | z, H_T, H_C, Y_C)  \\
&+ \log \frac{q(z|H_C, Y_C)}{q(z|H_T,Y_T,H_C,Y_C)}]
\end{aligned}
\label{elbo_1}
\end{equation}
where $q(z|H_T,Y_T,H_C,Y_C)$ is represented as $q(z|s_T)$ in the latent path, $q(z|HC,YC)$ is represented as $q(z|s_C)$, and $q(z|s_{\emptyset}):=p(z)$ is the prior on $z$. Then, the equation can be rewritten as

\begin{equation}
\begin{aligned}
\log p(Y_T | X_T,& X_C, Y_C)\\
\geq \mathbb{E}_{q(z|s_T)}  [ & \log p(Y_T | z, H_T, H_C, Y_C) + 
 \log \frac{q(z|s_C)}{q(z|s_T)}] \\
=\mathbb{E}_{q(z|s_T)}[& \log p(Y_T|z, H_T,H_C,Y_C)] \\
&- D_{KL}(q(z|s_T)||q(z|s_C))
\end{aligned}
\label{elbo_2}
\end{equation}
where KL is the Kullback-Leibler Divergence. Moreover, the representation of attention along the deterministic path is applied as $r^*_C=(H_C,Y_C,H_T)$. Therefore, the ELBO is derived as

\begin{equation}
\begin{aligned}
\log p(Y_T | X_T,  X_C, Y_C) & \geq\\
\mathbb{E}_{q(z|s_T)}[ & \log p(Y_T|z, H_T,r^*_C)] \\
&- D_{KL}(q(z|s_T)||q(z|s_C))
\end{aligned}
\label{elbo_3}
\end{equation}
As the learning and inference of ARNP model is completely achieved by a probabilistic treatment. The parameters of ARNP are learned by minimizing negative log-likelihood (NLL) following the reparametrization trick \cite{kingma2013auto_VAE}. Especially, the KL term encourages the summary of observed \textit{context} to be close to the summary of the \textit{targets} under the assumption that the \textit{contexts} and \textit{targets} come from the same realization of the data-generating stochastic process\cite{kim2019attentive_ANP}. Also, the predictions are expressed as the distribution of \textit{targets} value conditioned on observed \textit{contexts}. 

In a nutshell, the ARNP model incorporates Attentive Neural Process with RNNs explicitly, it is able to represent and quantify \textit{predictive uncertainty} in sequential data. The input sequence is first transferred into a latent space by a recurrent neural network, then the distribution over the functions of mapping the latent sequential information to the output is learned by a neural process enhanced by attention mechanism. 

\section{Experimental Results and Analysis}

\begin{figure}[t]
      \centering
      \includegraphics[width=\linewidth]{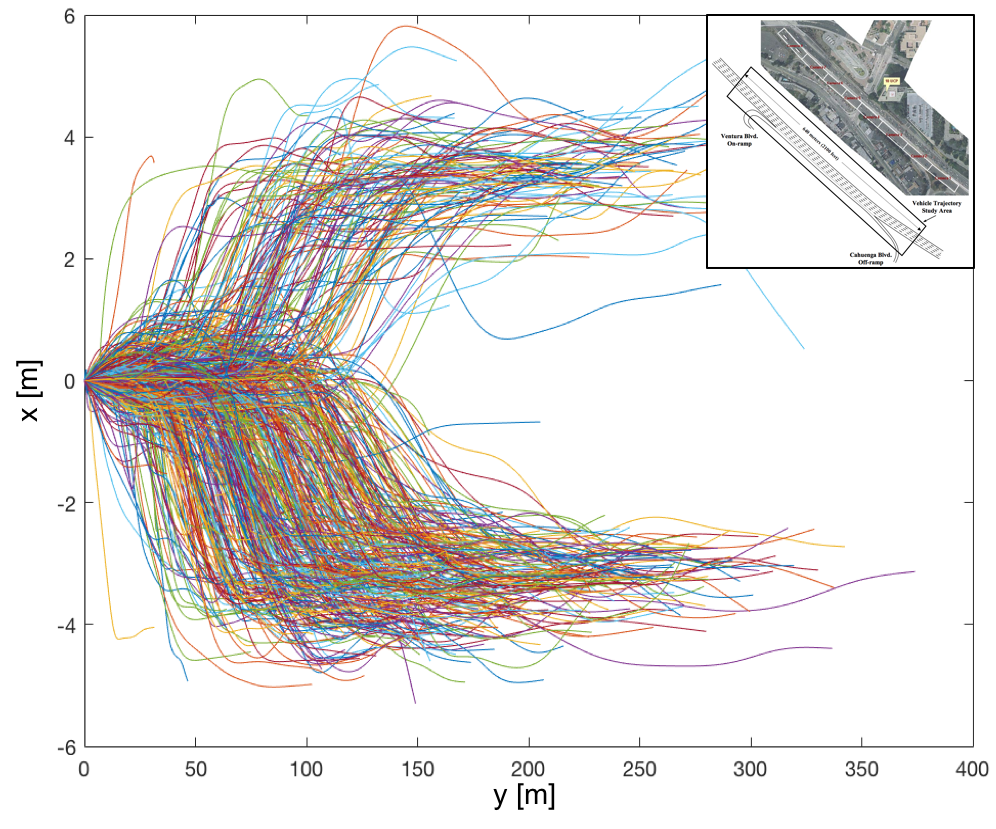}
      \caption{The visualization of 400 lane-changing trajectories collected from the NGSIM dataset.}
      \label{data}
\end{figure}


\begin{figure*}[t!]
    \centering
    \includegraphics[width=0.99\linewidth]{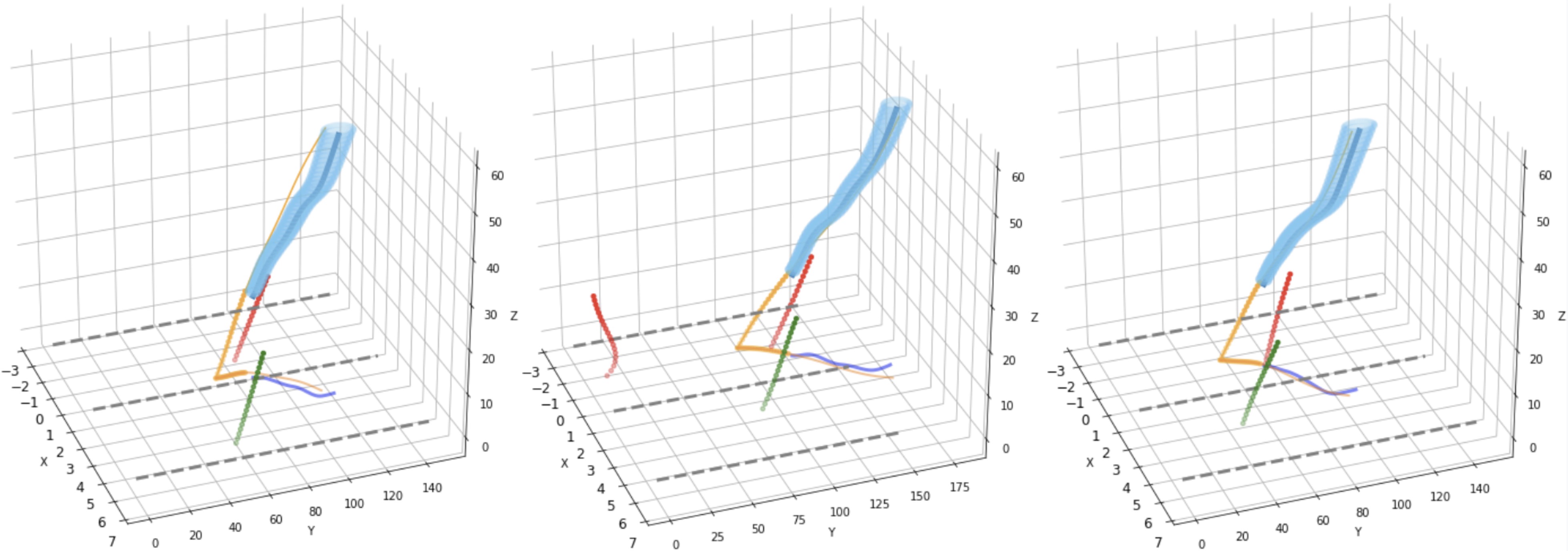}\\
    
    \text{(a)}
    \includegraphics[width=0.99\linewidth]{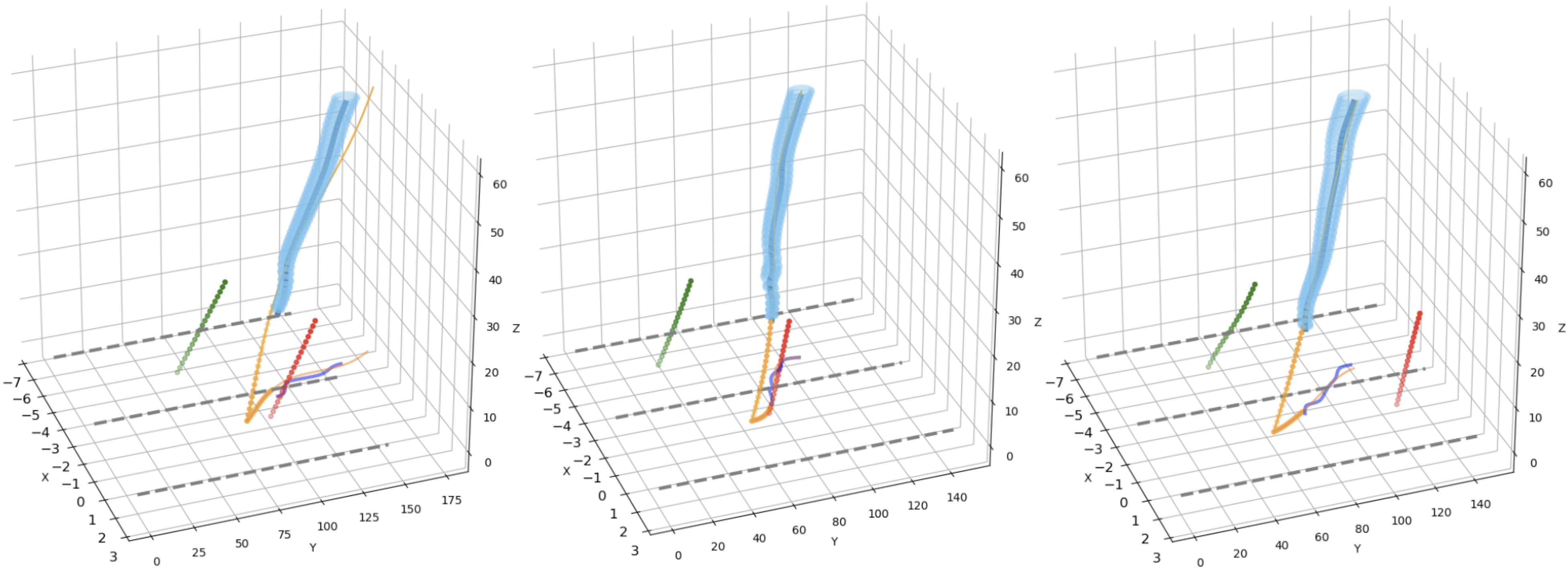}
    
    \text{(b)}
    \label{fig2b}
    \hfil
    \caption{Qualitative examples for (a) right lane-changing and (b) left lane-changing trajectory prediction of target vehicles. $x$-axis, $y$-axis, $z$-axis are the lateral direction, longitudinal direction, and the time horizon, respectively. The orange curves refer to the observed \textit{Ego Vehicle} trajectory and the slim orange curves refer to the ground truth. Black curves refer to the predictive mean of target vehicle trajectory, where the blue surface is the standard deviation. The blue curves on $x-y$ plane are the projection of predictive mean of trajectories. The red curves and green curves are the observed surrounding vehicle trajectories on the initial and target lanes.}
    \label{fig_lane_change}
\end{figure*}

\subsection{Experiment Description}

In order to train and test our proposed ARNP model for trajectory prediction tasks. Lane-changing scenarios were extracted from the NGSIM \cite{montanino2013making_ngsim} dataset from the I80 and the US101 highways, as shown in Fig. \ref{data}. The vehicle trajectory data contains precise location of each vehicle in the study area every one-tenth of a second.\par
Considering that when changing lanes, the driver usually only focuses on the vehicle on its origin lane and the target lane. In each scenario, except the \textit{Ego vehicle}, there are at most five surrounding vehicles, including the \textit{Front Vehicle}, the \textit{Following Vehicle}, and on the target lane, there are also \textit{Target Front Vehicle}, \textit{Target Immediate Vehicle} and \textit{Target Following Vehicle}, as shown in Fig. \ref{lane_changing_scenario}. The trajectories of all involved vehicles were recorded 10 seconds ahead to 10 seconds after the \textit{Ego Vehicle} crosses the lane-marking \cite{dong2017lane_gitdata}. 

\subsection{Baseline Models}

To demonstrate the improvements and features of ARNP in vehicle trajectory prediction, some baseline algorithms are implemented to compare with our proposed model. Regarding the problem defined as in section \ref{section3}, all the models predict the trajectory with the mutual interactions between the target vehicle and their surrounding vehicles. In another word, the historical trajectories of all involved vehicles are considered. We compare our proposed model with four baseline models described as follows.

\subsubsection{Neural Process (NP)} Neural Process \cite{garnelo2018neural_NP} is implemented as a baseline model for this trajectory prediction problem. NP also models a stochastic process that maps a set of inputs $X \in \mathbb{R}^{d_x}$ to a set of random variables $Y \in \mathbb{R}^{d_y}$. NP also uses a global latent variable $z$ to capture the global property of all training data. However, the prediction of the \textit{target} outputs are \textbf{permutation invariance}. This property works for applications such as image reconstruction and Bayesian Optimization \cite{garnelo2018neural_NP}, however it does not satisfy the consistency in learning sequential information. Specifically, the permutation invariance is obtained by aggregating local representation $\boldsymbol{r}_i$ for each observed input-output pairs by a a mean operator, $\boldsymbol{r} = \frac{1}{n} \sum^n_{i=1} \boldsymbol{r}_i$, and its dimension is 128. Similar as the structure in Fig. \ref{framework}, the representation $\boldsymbol{r}$ is obtained by a three-layer multilayer perceptron (MLP), the dimension of which is 32, 64, and 128. The latent representation $z$ is also computed by a MLP of the same structure. After obtaining the $\boldsymbol{r}$ and $z$, they are concatenated with \textit{target} input $X_T$ and fed into a generator network parameterized by $\theta$. 

\subsubsection{Attentive Neural Process (ANP)}
ANP is a generalization of NP, in which the attention mechanism is incorporated. In detail, in order to model the interactions between the \textit{context} points, the mean-aggregation of the representation of  \textit{context} input-output pairs $\boldsymbol{r}$ is replaced by a \textit{cross-attention}\cite{kim2019attentive_ANP}, where each \textit{target} input in $X_C$ attends to the observed \textit{context} to produce a query-specific representation. The ANP significantly improve the accuracy of NP and outperforms NP in 1-D function regression toy examples; however, it keeps the permutation invariance and is not suitable for sequential data.

\subsubsection{Recurrent Neural Networks (RNNs)}
As a special form of RNNs, Long Short-Term Memory (LSTM) \cite{hochreiter1997long_lstm} is one of the most successful methods for exploiting sequential information from data. LSTM takes input-output pairs incrementally in order, and learns long-term dependencies along the sequence by passing information in hidden units. Nevertheless, a plain LSTM learns the deterministic hidden information, thus can not model the propagating uncertainty in a sequence. Moreover, LSTM usually requires the input and prediction to be discretized.

\subsubsection{Reproducing Kernel Hilbert Space (RKHS)}
As an nonparametric approach \cite{Chiyu2018_itsc_rkhs}, RKHS provides a feasible and theoretically sound way to construct regression for trajectory planning tasks. It provides the prediction of continuous trajectories, and can also model the uncertainty by predicting a distribution. However, the application of RKHS faces obstacles due to its limited scalability.  

\subsubsection{Recurrent Meta Induction Neural Network (RMIN)}
The RMIN framework \cite{Chiyu2019_cnp_traj} pushes forward the progress of predicting interactive vehicle traejctory. This framework extracts the sequential information of the observed surrounding vehicles, and then aggregate the conditions into a Condition Neural Process (CNP). This model displays the capability of predicting vehicle trajectories. However, the RMIN framework neglected NPs' charming property of modeling stochastic process by only taking the CNP as a generator, and is not able to estimate the probability of vehicle behaviors explicitly.

\subsection{Results and Analysis}

\begin{table}[t]
  \centering
    \caption{Absolute Mean Errors and Standard Deviation of the Trajectory Prediction Compared with the Ground-True $\boldsymbol{h}$ in the Lateral Direction. The Unit is in Meters.}
  \begin{tabular}[b]{lllllll}
    \toprule
    \cmidrule(r){1-2}
         & \text{ } 
         & 1s 
         & 2s 
         & 3s
         & 4s
         & NLL\\
    \midrule
    
    \multirow{2}*{LSTM} & $\mu$  & 0.286 & -0.330 & -0.588 & -0.776 & ---\\
                        & $\sigma$ & 0.776 & 0.880 & 0.919 & 1.020 & --- \\[0.2cm]
    \multirow{2}*{RKHS} & $\mu$ & 0.052 & 0.251 & --- & --- & ---  \\
                        & $\sigma$ & \textbf{0.051} & \textbf{0.250} & --- & --- & ---\\[0.2cm]
    \multirow{2}*{RMIN}    & $\mu$ & \textbf{0.019} & \textbf{0.090} & 0.195 & 0.235 &---\\
                           & $\sigma$ & 0.501 & 1.299 & 1.997 & 2.343 & --- \\[0.2cm]
    \multirow{2}*{ANP}    & $\mu$ & --- & --- & --- & --- & 7024.45\\
                           & $\sigma$ & --- & --- & --- & --- & --- \\[0.2cm]
    \multirow{2}*{ARNP} & $\mu$ & 0.020 & 0.109 & \textbf{0.130} & \textbf{0.203} & \textbf{-0.0229} \\
                       & $\sigma$ & 0.235 & 0.276 & \textbf{0.307} & \textbf{0.332} & --- \\
    \bottomrule
  \end{tabular}
  \label{sample-table}
\end{table}


The prediction performance is evaluated based on the prediction error, sampled deviation, and negative log-likelihood (NLL), where the prediction error is computed by the Euclidean distance between the predicted trajectory and the real measured trajectory over time. The prediction performance of different methods is shown in Table \ref{sample-table}. Some methods are  unable to predict the distribution of trajectory directly, and the sample distribution is adopted after obtaining multiple predictive samples. Here, only the ANP and ARNP is trained by minimizing negative log likelihood (NLL) in a probabilistic perspective, while the other methods are either trained via optimizing with respect to a fine-tuned Mean Square Error (MSE) loss or solving a closed form function. \par 

Experiment results demonstrate that our proposed ARNP model achieves the lowest mean and variance for a 3 seconds prediction herizon, indicating its strong capability of capturing long-term dependencies among sequential data. Despite RKHS obtains the lowest variance in the first 2 seconds, it can only predict over a 2 seconds horizon \cite{Chiyu2018_itsc_rkhs}. The variance of ARNP is still comparable to RKHS. The the mean error of RMIN is slightly lower that ARNP by around $10\%$. In addition, given the fact that the RMIN is trained via minimizing the MSE loss while our proposed ARNP is trained by aligning the distributions of observed \textit{context} and \textit{targets} together. ARNP surely better expresses the underlying knowledge of these lane-changing behaviors. \par
Fig. \ref{fig_lane_change} shows a set of examples, where the target vehicle's trajectory is marked as thick orange lines, the surrounding vehicles are depicted as thick red and green lines, and the ground truth are slim orange lines. The mean and variance of predicted trajectory distributions are shown as black lines and blue surfaces. In order to provide a clearer prediction, the ground truth and predicted mean trajectories are projected to the ground plane, where dashed grey lines are the lane markers. As shown in the examples, our ARNP model not only predicts the trajectory accurately, but also products explicit distributions of trajectories. \par
As we can see, the ground truth trajectories are most captured by the predicted distributions that trend to accelerate after crossing the lane-markers, indicating that the ARNP captures a global trend from all the data.

\section{Conclusion}

This paper proposed the Attentive Recurrent Neural Process (ARNP) to generically learn traffic behaviors with the assumption that the different observed data from a traffic scenario are multiple realizations from a stochastic process. Our proposed model effectively capture the sequential information and the global underlying property of a traffic scenario. In a real-world trajectory prediction task, ARNP outperforms a number previous approaches in the aspect of prediction accuracy, variance, and probability expressiveness by predicting a explicit distribution. 
Our future work will follow the framework of neural processes and exploit the application of meta few-shoot learning by using ARNP model to recognize poorly understood traffic scenarios. 






\section*{Acknowledgement}
Toyota  Research  Institute  (“TRI”)  provided  funds  to  assist the  authors  with  their  research  but  this  article  solely  reflects the opinions and conclusions of its authors and not TRI or any other Toyota entity. \\
Thanks to Dr. Chiyu Dong for early discussions and suggestions.

\bibliographystyle{IEEEtran}
\bibliography{reference}

\end{document}